\documentclass[11pt,a4paper]{article}
\usepackage{times,latexsym}
\usepackage{url}
\usepackage[T1]{fontenc}
\usepackage{graphicx}
\usepackage[normalem]{ulem}

\usepackage[acceptedWithA]{tacl2021v1}

\usepackage{xspace,mfirstuc,tabulary}

\newcommand{\wctt}[1]{{\texttt{\footnotesize{#1}}}}

\newif\iftaclinstructions
\taclinstructionsfalse 
\iftaclinstructions
\renewcommand{\confidential}{}
\renewcommand{\anonsubtext}{(No author info supplied here, for consistency with
TACL-submission anonymization requirements)}
\newcommand{\instr}
\fi

\iftaclpubformat 

\else

\fi

\title{Watch Your Steps:\\Observable and Modular Chains of Thought}

\author{
  Cassandra A. Cohen
  \\
  \texttt{cassie.a.cohen@gmail.com}
  \And
  William W. Cohen 
  \\
  Machine Learning Dept
  \\
  Carnegie Mellon University\\
  \texttt{wcohen@cmu.edu}
}

\newcommand{\ourgh}{\url{https://github.com/wwcohen/doctest-prompting}}

\date{}

\newcommand{\ptp}{Program Trace Prompting}

\begin{document}
\maketitle
\begin{abstract}
We propose a variant of chain of thought (CoT) prompting called \ptp{} that makes explanations more observable while preserving the power, generality and flexibility of CoT.  In our approach, few-shot CoT demonstrations are wrapped in a formal syntax based on Python, and each prompt: identifies and names steps; defines the input/output behavior of steps; and replaces CoT explanations of in-context examples with chains of these formalized steps on the same examples.  \ptp{} is applicable to many tasks, achieving strong results on the 23 diverse tasks in the BIG-Bench Hard benchmark. More importantly, by instrumenting explanations in this way, we enable new types of analysis. In particular, we identify ``non-local errors'' (which correspond to incorrectly learning the reasoning method illustrated in the demonstrations) as an unaddressed issue in CoT learning, and we present methods for verifying the ``modularity'' of steps in a CoT explanation.
\end{abstract}

\iftaclpubformat
\section{Introduction}

While chain of thought (CoT) prompting is powerful, standard CoT outputs can be ``unfaithful'' \cite{jacovi-goldberg-2020-towards}: i.e., CoT can lead to incorrect but superficially plausible explanations for biased outputs \cite{turpin2024language}, and  CoT explanations ``may not align with \ldots sequential causal reasoning'' \cite{bao2024llms}.  

Although unfaithful explanations do not affect CoT's utility as a means of improving performance of prompted models, they do reduce the potential of CoT for other purposes, e.g., justifying a response to an end user.  Despite numerous proposals \cite{lanham2023measuring,bentham2024chainofthoughtunfaithfulnessdisguisedaccuracy,parcalabescu2024measuringfaithfulnessselfconsistencynatural}, unfaithfulness remains difficult to detect and measure. The potential unfaithfulness of CoT explanations presents a jarring contrast with symbolic proofs, in which modular, verifiable reasoning steps are combined in well-understood ways.  

We believe that a significant obstacle to progress on the faithfulness of CoT prompts is their syntactic diversity; because a CoT explanation can take nearly any form, they are difficult to analyze in any general way.
The goal of this paper is \emph{to make CoT explanations easier to analyze while preserving the power, generality and flexibility of CoT}.  To do this, we propose a new variant of CoT prompting in which few-shot CoT demonstrations are wrapped in a semi-formal syntax which (1) identifies and names steps; (2) defines the input/output behavior of steps; and (3) replaces each CoT explanation in a demonstration with an equivalent chain of formalized steps.  

\begin{figure*}[htb]

\centering
\includegraphics[width=0.9\textwidth]{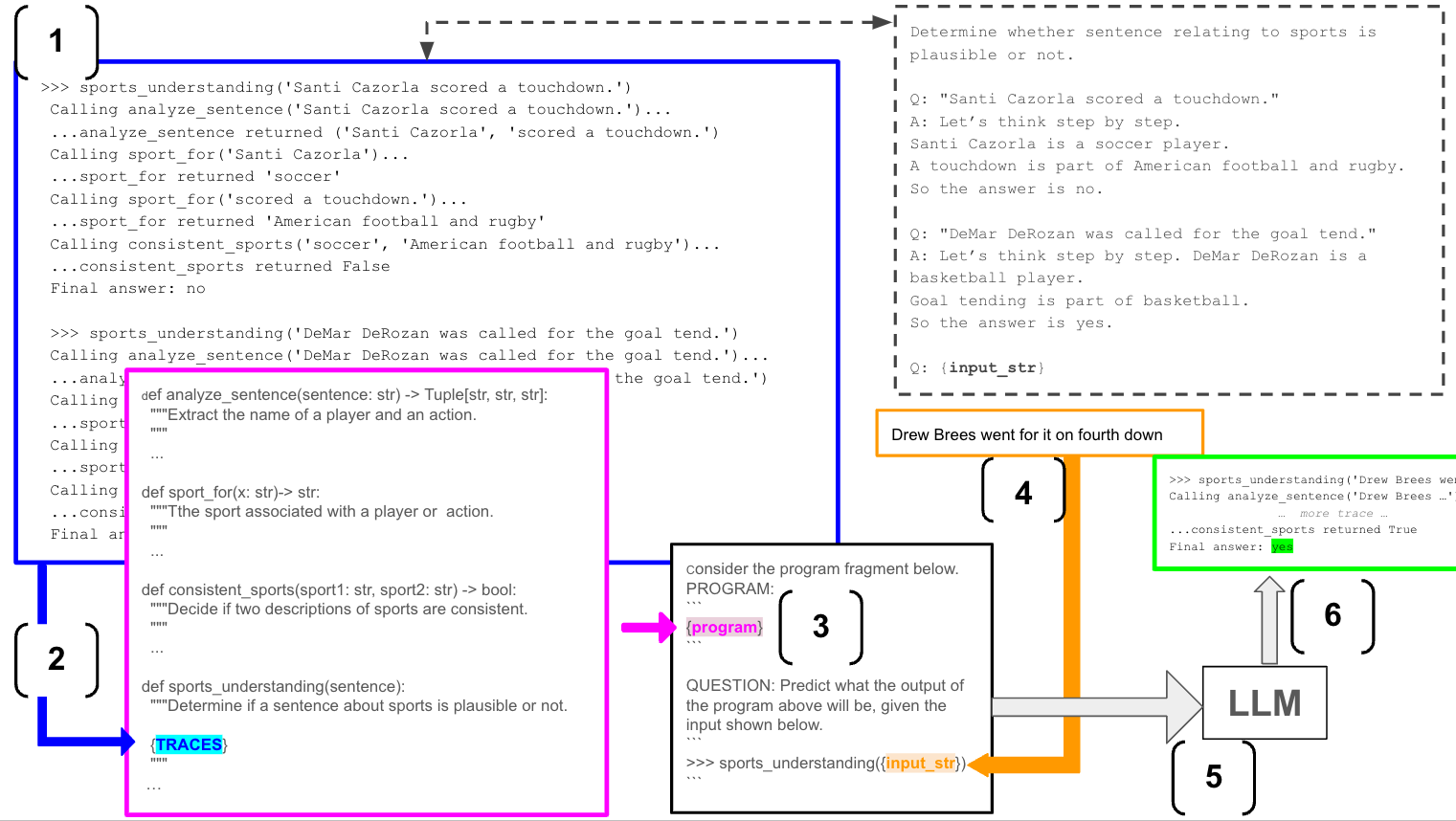}

\caption{An illustration of \ptp{} on a simplified version of a task from Big Bench Hard.  (1) Instead of the original CoT prompt, we begin with a trace of a Python pseudo-program that implements a similar problem-solving strategy (see text for how this trace is generated).  (2) This trace is then inserted into a set of ``stubs'', which document the subroutines used in the trace, yielding a skeleton of a program, which contains traces, type signatures, and documentation, but no code.  (3) The skeleton program is inserted, along with a test input (4), into a prompt that instructs a LLM to predict the output of the program on the test input. 
(5) This prompt is then sent to an LLM, which produces (6) a predicted program trace, which contains the desired prediction for the test output--in this case, the word ``yes''.}
\label{fig:fig1}
\end{figure*}

We use Python syntax to describe steps, and call our method \ptp{} (PTP).  As shown in Figure~\ref{fig:fig1}, CoT demonstrations are replaced with documentation for a Python program, together with traces of that program's behavior on the demonstration inputs.  Each different kind of ``step'' is associated with a Python subroutine, and in addition to traces, the prompt includes \emph{stubs} (function-level comments and type signatures) for these subroutines. An LLM is then prompted with this information as context, and asked to produce a trace for a novel input. 
An answer to the question is finally extracted from the LLM-generated trace.

Note that in \ptp{}, \emph{no code is presented to the LLM}. 
Similarly,  there is no ``tool use''; no generation of code or pseudo-code to be executed by Python or some other engine; and no pipeline of LLM calls as in agent frameworks like LangChain \cite{topsakal2023creating} or DsPy \cite{khattab2023dspy}. Instead, for a new test input, the LLM simply generates a new trace, similar to the ones given in the prompt, but appropriate for the test input.  

Importantly, the program suggested by the traces typically do not exist, and in the more interesting applications of \ptp{}, \emph{cannot} exist as pure Python.  In Figure~\ref{fig:fig1}, for example, the subroutine \wctt{sport\_for} requires using background knowledge about arbitrary entities and phrases.  While this cannot be implemented directly in Python, a strong LLM can readily perform the underlying task 
and thus generate a trace.  Hence, the traces generated by PTP are best thought of as traces of Python-like ``pseudocode'' \cite{li2023chain,weir2024learning}.

In this paper, we describe a particular implementation of \ptp, and demonstrate that it has the following properties.

    First, \emph{PTP is applicable to a broad range of tasks.}  We implement PTP prompts for each of the 23 diverse tasks in BIG-Bench Hard (BBH) \cite{suzgun2022challenging}, and show that these have accuracy comparable to the corresponding CoT prompts.
    
    Second, \emph{PTP outputs can almost always be automatically parsed into legal sequences of the defined steps.} 
    In generated traces for unseen test cases, there were \emph{no} hallucinated subroutine names, and more 99\% of the ``steps'' (i.e., the function calls in traces) were well-formed.  
    For well-formed function calls, over 95\% of the calls are also syntactically correct Python objects of the expected type.

    Third, \emph{PT prompts can be used to execute individual steps, as well as solve complete instances of the task.}  For instance, by replacing the input \texttt{sports\_understanding('Drew Brees went for it on fourth down')} with \texttt{sport\_for('Drew Brees')}, we could use the prompt of Figure~\ref{fig:fig1} to generate only the first step of the PTP trace for this task.  Experiments show that this process has accuracy of over 90\%, averaged over 16 steps from 6 tasks.

    Fourth, \emph{PTP traces can be analyzed in ways conventional CoT explanations cannot.}
    In particular, we show that errors can be manually evaluated for \emph{locality}, and that the steps in PTP traces can additionally be automatically evaluated for \emph{modularity}.
    As defined here, a step is \emph{modular} if its behavior depends only on the information labeled as inputs to the step. For instance, the step \wctt{sport\_for(``scored a touchdown'')} is modular if the output depends only on the input string \wctt{``scored a touchdown''}, and is not influenced by other previously-generated text in the trace.  In Section~\ref{sec:modularity} we present definitions, and a protocol for automatically detecting non-modular steps.  Experimental results show that most steps in our collection of tasks are modular.

    Since steps in our model are  explicitly represented, one can characterize errors in reasoning as \emph{local} to a particular step. However, a final answer may be wrong even when every local step seems to be correct.  We call these \emph{non-local errors}, but a more descriptive name might be ``program induction errors''; intuitively, non-local errors arise because the LLM has ``guessed'' the wrong algorithm from the few-shot program traces. In our analysis we show that the majority of incorrect final answers are associated with local errors, especially for the algorithmically-simpler NLP tasks. 


To summarize, the contributions of this paper are (1) description of a new framework for CoT prompting that allows detailed analysis of the ``steps'' used in a CoT explanation; (2) demonstration of the generality of this framework by testing it on 23 diverse tasks, and releasing the corresponding prompts to the community;  (3) presentation of methods to evaluate two newly-defined aspects of CoT reasoning, namely the modularity of CoT steps and the locality of CoT errors; and (4) an analysis of the modularity of steps over a large set of tasks and step types. 

\section{Methods}

\subsection{Data}

\begin{table*}
\begin{footnotesize}
\begin{tabular}{rl}
\hline
Task & Description \\ \hline
\multicolumn{2}{l}{\textit{NLP Tasks}} \\
Causal Judgment & Answer questions about causal attribution \\
Date Understanding & Infer a date from context \\
Disambiguation QA & Clarify the meaning of sentences with ambiguous pronouns \\
Formal Fallacies Syllogisms Negation & Distinguish deductively valid arguments from fallacies \\
Hyperbaton (Adjective Ordering) & Order adjectives correctly in English sentences \\
Movie Recommendation & Recommend movies similar to a given list of movies \\
Penguins in a Table & Answer questions about a table of penguins and their attributes \\
Reasoning about Colored Objects & Answer questions about the colors of objects on a surface \\
Ruin Names & Select the humorous edit that 'ruins' a movie or musical artist name \\
Salient Translation Error Detection & Detect the type of error in an English translation of a German sentence \\
Snarks & Determine which of two sentences is sarcastic \\
Sports Understanding & Determine whether a sentence relating to sports is plausible or not \\ \hline
\multicolumn{2}{l}{\textit{Algorithmic Tasks}} \\
Boolean Expressions & Evaluate the result of a random Boolean expression \\
Dyck Languages & Correctly close a series of open/close parenthesis of different types \\
Geometric Shapes & Name geometric shapes from their SVG paths \\
Logical Deduction & Deduce the order of a sequence of objects given constraints \\
Multi-Step Arithmetic & Solve multi-step arithmetic problems \\
Navigate & Given navigation instructions, determine whether one ends up back at the starting point \\
Object Counting & Count the number of objects of different types \\
Temporal Sequences & Answer questions about which times certain events could have occurred \\
Tracking Shuffled Objects & Find the final positions of objects given initial positions and a sequence of swaps \\
Web of Lies & Evaluate a random boolean function expressed as a word problem \\
Word Sorting & Sort a list of words \\ \hline
\end{tabular}
\end{footnotesize}
\caption{Tasks in BIG Bench Hard.  For each task, there a 3-shot CoT prompt and 250 test examples (except for ``Snarks'', which has only 178 examples, and ``Causal Judgement'', which has only 187.) 
}
\label{tab:bbh}
\end{table*}

We used the Big-Bench hard \cite{suzgun2022challenging} tasks to evaluate our approach (see Table~\ref{tab:bbh}).  This is a well-studied and diverse set of 23 tasks which are known to be challenging for LLMs.  All the tasks have 3-shot CoT prompts, and all the tasks also well-suited to evaluation, having answers that can be easily tested for correctness (most of the tasks are multiple-choice.)
About half the tasks are considered ``NLP'' tasks, and are broadly similar to the simplified example of Figure~\ref{fig:fig1}: a high-level reasoning strategy is followed, which requires calling some low-level routines, and the low-level routines require some kind of AI.  The other half are considered ``algorithmic'' tasks, which can be solved with simple extraction rules and an algorithm.  For instance, one of the algorithmic tasks is evaluating a Boolean expression; another is reasoning with logical constraints on the ordering of a small number of objects.

\subsection{Constructing prompts}

\subsubsection{Mocks based on CoT prompts}
In all our experiments, we constructed program trace prompts that closely followed the existing CoT prompt for the task (as suggested by Figure~\ref{fig:fig1}.)  The existing CoT prompts were also used as a performance baseline.  Modifications to CoT prompts were made only (1) to encourage the model to closely follow the in-context demonstrations, and in particular to produce output in an easily-parsed format and (2) rarely, to fix reasoning errors in the CoT prompt.  See Appendix~\ref{app:cot} for details.

Program trace prompts were produced semi-automatically.  For each task, we manually wrote a Python program that implements an algorithm suggested by the CoT prompt; this is feasible because the program only needs to  for the few examples appearing in the CoT prompt.  (For instance, the implementation of \wctt{sport\_for} just retrieves the required values from a small dictionary  containing entries for the specific entities and phrases in the CoT prompt.) Following the parlance of unit testing, we call these programs \textit{mocks}.\footnote{Mocks are a standard way of testing interactions with a complex system that you don't want to actually call at testing time.}  Functions can be nested (recursively if necessary) in a mock.
When a mock is executed it automatically prints a trace; see Appendix~\ref{app:tracegen} for more information.  Our traces primarily record function calls, although they can also include printed output---notably, as in  Figure~\ref{fig:fig1},  the trace contains printed output for the ``final answer'' to a question.  

Our placement of the traces (in the function-level documentation of the top-level function) follows a widely used convention for documenting function behavior by giving input/output examples.\footnote{The Python3 \wctt{doctest} module supports unit testing for functions that follow this convention.} 

More details are in Appendix~\ref{app:tracegen}.

\subsubsection{Mock development}

Our process is semi-automatic because the mocks are written by hand. For many tasks, the mock is a thin wrapper around the steps in the CoT demonstrations; however, some meaningful decisions do need to be made by mock authors.  For instance, in Figure~\ref{fig:fig1}, the function \wctt{sport\_for} might plausibly return a list of strings, rather than a single string, and the \wctt{consistent\_sports} function could plausibly be omitted, as it is implicit in the CoT prompt.

To reduce overfitting when making these decisions, we used 60 random examples of each task to help design the mocks, and reserved the rest (190 examples for most tasks) as an unseen test set set.  The 60 examples are further divided into ``dev'' and ``tuning'' sets with 30 examples in each.  Mock implementation decisions were made based on dev data performance, but the actual values in the mocks are based on the CoT prompts, not the dev set.
The tuning set was held out in early mock design for the annotation experiments of Section~\ref{sec:error-types}.

It should be noted that this procedure is different from most common prior uses of BBH.  Usually, BBH is used to evaluate the performance of a new LLM, by testing its performance using the provided CoT prompts.  Although BBH has been
used for evaluating new prompting methods, there appears to be no standard dev/test split that is used to prevent developers of prompting methods from overfitting to the test data. Our splits and  experimental scripts
are available on GitHub\footnote{\ourgh{}}.

\subsection{Special prompting strategies}

\subsubsection{Prompting to perform a single step} \label{sec:step-by-step}

One interesting aspect of \ptp{} is that the same prompt can be used to either predict traces for either a full CoT inference process, or \emph{any individual step}, by just replacing the code fragment at the end of the prompt used in stage (3) of the \ptp{} process of Figure~\ref{fig:fig1}.  For example, to predict a trace for the step \wctt{sport\_for('scored a touchdown')}, one could replace the text 

\wctt{>{}>{}> sports\_understanding(\{input\_str\})}

\noindent in the stage (3) prompt with 

\wctt{>{}>{}> sport\_for('scored a touchdown')}

An ideal generation for the prompt above might be something like

\noindent \begin{scriptsize}
\begin{tt}
\begin{tabular}{l}
{Calling sport\_for('scored a touchdown')...}\\
{...sport\_for returned 'American football'}\\
{'American football'}
\end{tabular}
\end{tt}    
\end{scriptsize}

\noindent The final output of the step is then simply the last line of the generation.

To our knowledge, there is no counterpart to this kind of single-step execution in conventional CoT prompting, or any of its many variants---it is a consequence of the fact that we have made steps more explicit.  

One can use these single-step prompts to do something very much like interactive debugging of a mock, by exploring its behavior on subroutines.  We also use this for various analytic tasks in Section~\ref{sec:expts}.

\paragraph{Improvements to single-step prompting} \label{sec:trace-parsing}

The result of single-step prompting is more variable than the result of PTP prompting on full examples.  One common failure mode for  single-step prompts is that the LLM returns a trace that begins with the requested step, but then continues, usually tracing steps that might plausibly follow the requested step.  We call this \emph{overgeneration.} 

Overgeneration can be reduced by two methods. One is to add  single-step traces to the stubs for the subroutine being probed.  Empirically, we find that performance for  single-step prompting is improved even if the only traces added are ones that are part of the top-level traces.\footnote{Specifically, improved performance is obtained by just copying over the appropriate part of the top-level CoT trace into the ``stub'' defining a step, as discussed in Section~\ref{app:single-step}.}  A second method is to use more a robust method for extracting output from  single-step generations, i.e., parsing the trace to identify steps, and then using, as the  single-step result, the return value from the first step with correct subroutine name.  More details are given in Section~\ref{sec:modularity} and Appendix \ref{app:single-step}.


\subsubsection{Completing a partial trace} \label{sec:complete-trace}

A similar prompting trick can be used to complete a trace given a prefix of the trace: we just replace the question in the stage (3) prompt with one that requests the LM to complete the trace, rather than generate it. See Appendix~\ref{app:continue-trace} for details.

\subsection{Language models}
In principal, \ptp{} can be used with any LLM.  After preliminary studies, we used Anthropic's Sonnet 3 model (2024-06-20) as the basis for developing our prompts, as it obtains a good balance between cost, reliability, and performance. Unless we state otherwise, this is the LLM model used in the experiments below.  \ptp{} was also tested on other models, as described below.

\section{Results: Accuracy and Types of Errors} \label{sec:expts}

\subsection{How well do PT prompts work?}

\newcommand{\B}[1]{\textbf{#1}}

\begin{table}[t]
\begin{small}
\begin{tabular}{lrcr}
\hline
                        &        PTP 	&       & CoT	\\
\hline
Causal Judgement &		\B{65.8}&		& 64.9 \\
Date Understanding$*$&		88.8&		& \B{92.6} \\
Disambiguation QA&		\B{82.6}&		& 80.0 \\
Formal Fallacies&		56.8&		& \B{57.4} \\
Hyperbaton&	        	\B{97.9}&	$\gg$	& 85.8 \\
Movie Recommendation&		\B{91.1}&		& 87.9 \\
Penguins in a Table&		89.2&		& \B{92.8} \\
Reasoning Colored Objs&		94.2&		& \B{94.7} \\
Ruin Names&	        	83.2&	&	 \B{86.6} \\
Salient Translation Errs&		69.5&		& \B{70.5} \\
Snarks&	                	\B{91.5}&	& 89.0 \\
Sports Understanding&		\B{97.4}&		& \B{97.4} \\
\hline
Average NLP             &	\B{84.0}    &	& 83.3 \\
\hline
Boolean Expressions&		\B{95.3}&		& 92.6 \\
Dyck Languages&	        	\B{76.2}&	$>$	& 62.7 \\
Geometric Shapes&		40.0& $\ll$	&	 \B{53.3} \\
Logical Deduction       &	87.9    & $\ll$  & \B{96.3}   \\
Multistep Arithmetic&		\B{87.9}&	$\gg$	& 73.7 \\
Navigate$*$&	        	\B{98.9}&		& 97.9 \\
Object Counting&		\B{100.0}&	&	 98.9 \\
Temporal Sequences$*$&		96.8&		& \B{100.0} \\
Tracking Shuffled Objs&		98.9&		& \B{100.0} \\
Web of Lies&	        	\B{100.0}&		& \B{100.0} \\
Word Sorting&	        	\B{96.3}&		& 91.4 \\
\hline
Average Algorithmic     &	\B{88.9}&		& 87.9 \\
\hline
\hline
Average&	        	\B{86.4} &		& 85.5 \\
\hline
\end{tabular}
\end{small}
\caption{Accuracy on BBH problems, using a test set unseen during development of PTP prompts. All prompts were run using Anthropic Sonnet 3, the LLM used for PT prompt development. Statistically significant differences are shown in the third column, with $p<5\%$ indicated by $<$ or $>$, and $p<1\%$ by $\ll$ or $\gg$. 
} 
\label{tab:ptp-vs-cot}
\end{table}

Our goal is to instrument CoT prompts so that they can be more easily analyzed---but we would like to do that without impacting the performance.  Hence, we first look at the overall accuracy of PTP prompts compared to the CoT prompts on which they are based.

Table~\ref{tab:ptp-vs-cot} summarizes the accuracy results on the BBH tasks.  To avoid prompt-tuning that might overfit the test data, the results in the table are generally\footnote{For the three starred cases, problems discovered were in the PTP prompt late in experimental procedure, and the results are for the second prompt tested--see Appendix~\ref{app:expts}.} for the \emph{first} PT prompt for a task that was evaluated on the test data.

The results of Table~\ref{tab:ptp-vs-cot} strongly support our first claim, that \emph{PTP is applicable to a broad range of tasks}, in the sense that it can be used on a broad range of tasks without a major impact in performance.  Of the 23 diverse tasks in BBH, switching from an informal format for explanations to a more precise one led to a statistically significant loss in performance for only two tasks.  Three tasks showed significant improvements in performance, and average performance of PT prompts is slightly improved over the CoT baselines.

As discussed above, our experimental procedure differs from most prior work, but the CoT baseline we use is strong compared to many prior papers; see Table~\ref{tab:prior} in the Appendix.

In Table~\ref{tab:other-models}, we also show results comparing PTP and CoT prompting for several other proprietary models, one strong open-weight model (Deepseek Coder V2, a 236B model), and several smaller models.  These experiments were run on a subset of the data, namely the dev set for six tasks chosen to represent different kinds of tasks and different levels of accuracy.  For most strong models the average performance is similar (within 3-4 points) for both prompting methods; however, PTP underperforms CoT prompting on many smaller models.

\begin{table}
\centering
\begin{tabular}{ccc}
Model & PTP & CoT \\
\hline
\multicolumn{3}{c}{\textit{Commercial models}}\\
\hline
Anthropic Sonnet-3.5 & 92.8	& 92.2 \\
Anthropic Sonnet-3   & 85.0	& 83.9 \\
Anthropic Haiku	     & 73.9	& 75.0 \\
\hline
Gemini Pro-1.5	     & 90.6	& 86.1 \\
Gemini Flash-1.5     & 84.8	& 92.0 \\
\hline
OpenAI GPT4o	     & 87.2	& 90.6\\
OpenAI GPT4o-Mini    & 68.0	& 92.6 \\
\hline
\multicolumn{3}{c}{\textit{Open-Source models}}\\
\hline
Deepseek Coder-v2    & 89.2	& 92.0 \\
\hline
Llama-3.1 7B         & 64.5 &	77.8\\
CodeLlama 7B         & 41.2 & 50.5 \\
\hline
Gemma2 9B             & 69.3 & 82.2 \\
Gemma2 2B            & 31.0 & 43.8 \\
\hline
\end{tabular}


\caption{Average results on other models.  All results are over the dev set for a subset of six tasks (Date Understanding, Hyperbaton, Sports Understanding, Geometric Shapes, Multistep Arithmetic Two, Logical Deduction Three Objects).} \label{tab:other-models}
\end{table}

\subsection{Are generated traces syntactically well-formed?}
\label{sec:autoparse}

\begin{figure*}[htb]

\centering
\includegraphics[width=0.5\textwidth]{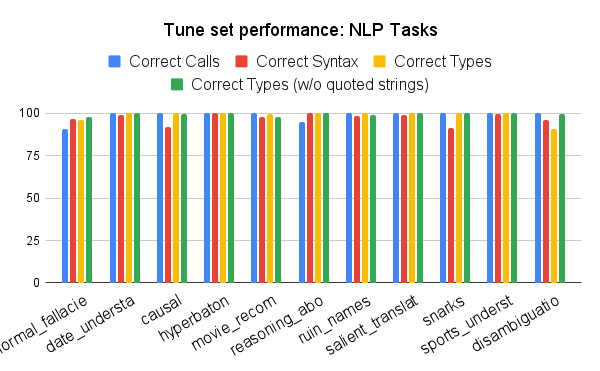}\includegraphics[width=0.5\textwidth]{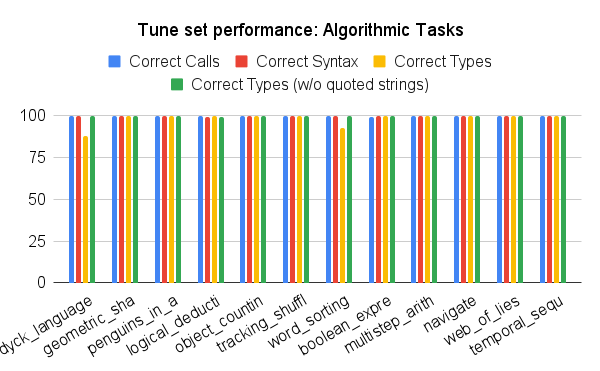}

\caption{Syntactic well-formedness of generated traces BBH tasks.}
\label{fig:tune-analysis}
\end{figure*}

\begin{table}[tb]
\begin{center}
\begin{tabular}{lr}
\hline
                  & Avg \\
\hline
Well-formed calls & 99.3\%     \\
\hline
No syntax errors  & 98.6\%   \\
~~ignoring str errors  & 99.7\%   \\
No type errors    & 98.6\%    \\
\hline
\end{tabular}
\end{center}
\caption{Overall correctness of calls, argument/return value syntax, and argument/return value types on the tune set for 23 BBH tasks.}  \label{tab:tune-analysis}
\end{table}

The main goal of \ptp{} is to produce structured explanations, and thus enable deeper analysis of CoT performance.  To verify that the explanations do indeed have the intended structure, we wrote tools to automatically evaluate generated traces for the following properties.

    \textit{Call correctness.} Traces were parsed to see if they either (a) contain ``hallucinated'' calls to functions not listed in the prompt or (b) are a well-formed function-call trace (i.e., every \wctt{"Calling f(a, b)..."} line can be paired with a corresponding \wctt{"...f returns c"}  line, following nesting rules.)   
    
    \textit{Syntactic correctness.}  For well-formed traces, the arguments and return value for every function call were checked for syntactic correctness according to Python's rules (i.e., whether the LLM outputs could be evaluated to construct a Python object).  
    
    \textit{Type correctness.}  Syntactically correct arguments and return values were checked to see if they were type-correct, according to the type hints given in the stubs for each step.

The analyses of this paper only require call correctness, but interestingly, \ptp{} generally produces traces that are syntax- and type-correct as well.  Figure~\ref{fig:tune-analysis} gives a detailed breakdown of this analysis (on the tune set) for each BBH task, and Table~\ref{tab:tune-analysis} gives a summary average over all tasks. 
The vast majority of syntactic errors, where a value passed to or returned from a step are not legal Python objects, are associated with strings that contain characters (e.g., quotation marks) that have not been escaped correctly; the table line ``ignoring string errors'' gives the accuracy if these errors were ignored.

This analysis supports our second claim, that \emph{the traces generated by PT prompts can almost always be automatically parsed into legal sequences of the 
predefined steps associated with the prompt.}  This  allows us to further analyze the traces and characterize their errors.

\subsection{What kinds of errors are made?} \label{sec:error-types}

\begin{figure}[tb]
\begin{scriptsize}
    \begin{tabular}{l}
\hline
Calling sport\_for('Santi Cazorla')...\\
...sport\_for returned 'soccer'\\
Calling sport\_for('scored a touchdown.')...\\
...sport\_for returned 'NFL football and rugby'\\
Calling consistent\_sports('soccer', 'NFL football and rugby')...\\
...consistent\_sports returned False\\
\hline
Calling sport\_for('Santi Cazorla')...\\
...sport\_for returned \textcolor{red}{'rugby'}\\
Calling sport\_for('scored a touchdown.')...\\
...sport\_for returned 'NFL football and rugby'\\
Calling consistent\_sports(\textcolor{red}{'rugby'}, 'NFL football and rugby')...\\
...consistent\_sports returned \textcolor{red}{True}\\
\hline
Calling sport\_for('Santi Cazorla')...\\
...sport\_for returned \textbf{\textcolor{red}{'soccer'}}\\
Calling sport\_for('scored a touchdown.')...\\
...sport\_for returned \textbf{'NFL football and rugby'}\\
Calling consistent\_sports(\textbf{\textcolor{red}{'rugby'}, 'NFL football and rugby'})...\\
...consistent\_sports returned \textcolor{red}{True}\\
\hline
\end{tabular}
\end{scriptsize}
\caption{Top, part of a correct program trace.  Middle,
a \emph{local} error: the first \texttt{sport\_for} step returns an incorrect result (red), which causes an incorrect answer.  Bottom, a \emph{non-local} error: the \texttt{consistent\_sports} call should have the bold-faced arguments 1 and 2 copied over from the first and second \texttt{sport\_for} outputs, respectively, but the red bold-faced argument was not copies correctly.} \label{tab:error-type}
\end{figure}

Since the PTP prompt specifies the semantics of individual steps, is possible to annotate \emph{where} errors occur. In general, errors in PTP reasoning can occur either because a step produces a wrong output (which we call here a \emph{local error}),  or because a wrong sequence of steps is used (which we call a \emph{non-local error}). Examples of these error types are illustrated in Figure~\ref{tab:error-type}; as suggested, the most common non-local error involves copying values from the wrong place.  

\begin{table}[tb]
\begin{small}
\centering
\begin{tabular}{lrr}
\hline
             & \% of Traces & \% of Errors \\
             \hline
 All errors   & 11.7  & 100.0 \\
 Local errors &    9.1    &  77.8 \\
 Non-local errors & 2.6 &  22.2 \\
\hline
\end{tabular}
\caption{Analysis of local-vs non-local errors in BBH hard tasks on the tuning set.}  \label{tab:nonlocal}
\end{small}
\end{table}

\begin{table}[tb]
\begin{footnotesize}
\begin{tabular}{lrrr}
\hline
Task & Avg \# & Trace & Non-local \\
     & Steps & Entropy & Error Rate \\
\hline
object counting & 3.0 & 0.00 \\
web of lies & 6.0 & 0.00     \\
snarks & 7.0 & 0.00  \\
movie recommendation & 9.0 & 0.00    \\
tracking shuffled objects & 10.0 & 0.00 \\
logical deduction  & 10.0 & 0.03 \\
salient translation errors & 4.0 & 0.03     \\
temporal sequences & 4.0 & 0.03      \\
disambiguation qa & 7.0 & 0.11       \\
reasoning @ colored objs & 3.9 & 0.12 \\
sports understanding & 4.3 & 0.44    \\
multistep arithmetic two & 26.2 & 1.09  & 3.3\%     \\
ruin names & 18.4 & 1.14      \\
date understanding & 11.3 & 1.19  & 3.3\%   \\
penguins in a table & 3.6 & 1.20  & 3.3\%  \\
hyperbaton & 11.5 & 1.59      \\
causal judgement & 6.0 & 1.76       \\
geometric shapes & 17.1 & 1.76  & 13.3\%     \\
navigate & 7.3 & 1.98        \\
dyck languages & 42.7 & 4.00  & 23.3\% \\
formal fallacies & 12.3 & 4.15 & 13.3\%       \\
boolean expressions & 7.4 & 4.33     \\
word sorting & 47.8 & 4.83    \\
\hline
\end{tabular}
\end{footnotesize}
\caption{Number of steps and trace abstraction entropy of the 23 BBH Tasks (on the test data).  Lower trace entropy indicates a more predictable sequence of steps.
Only non-zero non-local error rates are shown.
} \label{tab:trace-ent}
\end{table}

We manually annotated all of the traces from the tune set that led to an incorrect final result. We marked the first\footnote{Only the first local error was marked, because errors often cascade in CoT reasoning chains.} clearly incorrect step output, if there was one, and otherwise marked the trace as containing a non-local error.  The results are shown in Table~\ref{tab:nonlocal}, with more detail on non-local errors in Table~\ref{tab:trace-ent}.

\subsection{Why do non-local errors occur?} \label{sec:why-nonlocal}

To our knowledge, the existence of non-local errors in CoT reasoning has not been discussed previously in the literature.  There are several possible reasons for this.  First, non-local errors are relatively rare:  only 2.6\% of the traces had non-local errors.  Second, most non-local errors occurred in a small number of tasks:  more than 80\% of the non-local errors were in three of the tasks (Dyck Languages, Geometric Shapes, or Formal Fallacies), so more than 85\% of the tasks contained very very few non-local errors.  Third, non-local errors are hard to detect, even by careful manual annotators.  Despite this, non-local errors are not unimportant: 22\% of the incorrect traces were annotated as having non-local errors.  

Our conjecture is that non-local errors are correlated with the difficulty of learning to sequence the steps: in other words, \emph{non-local errors arise when the algorithm suggested by the CoT trace is complex, and hence difficult to learn.}

\label{sec:abstract-trace}

To test this, we analyzed the traces produced for each task on the test examples. For every legal trace, we simplify the trace by removing the arguments and return values, leading to a \emph{abstract trace}.  For example, the traces of Figure~\ref{fig:fig1} all have the same abstract trace: \wctt{analyze\-\_input sport\-\_for sport\-\_for consist\-ent\-\_sports}.
For some tasks, only one abstract trace is ever generated on any test example, but many tasks have diverse abstract traces, especially tasks with inputs that vary in size or structure.  


To quantify the diversity of abstract traces, we computed the entropy of the empirical distribution of abstract traces for each task, which we show in Table~\ref{tab:trace-ent}.  
There is a clear correspondence between high trace entropy and non-local errors: the three highest non-local error rates are among the six highest-entropy tasks.  The correlation between error rate and trace entropy is $r=0.51$.  In contrast, the correlation to the number of steps in the trace (also shown in the table) is only $r=0.04$.

To summarize, \emph{structuring CoT explanations with \ptp{} not only reveals the existence of non-local errors, but also allows us to predict where non-local errors might occur.}  

These results also emphasize a little-discussed function of CoT prompting: in addition to ``unlocking reasoning'' \cite{wei2022chain}, a CoT prompt can also provide a new sort of information to the LLM, as its steps describe a particular \emph{strategy} for solving a task, which the LLM can then learn.  This ``strategy learning'' seems to be very reliable for tasks of low and medium complexity.

\section{Results: Modularity}

\subsection{Are individual steps meaningful?} \label{sec:modularity}

We now turn to the question of how reliably individual steps are learned.
The results of Section~\ref{sec:autoparse} show that PTP traces syntactically \emph{look} like sequences of well-defined steps, each of which has known inputs and outputs.  However, past work on evaluating the ``faithfulness' of CoT explanations \cite{turpin2024language,bao2024llms} shows that appearances of this sort  can be deceiving. To determine if the abstraction of a ``step'' is meaningful, also we evaluated how well LLMs can execute individual steps, using the prompting scheme of Section~\ref{sec:step-by-step}.  

\subsubsection{Individual steps can be evaluated in isolation} \label{sec:oracle}


Evaluating performance on a single step is difficult, since we do not have gold labels for each step's output. 
Fortunately, several individual steps in the BBH tasks are relatively easy to encode with rules---so easy, in fact, that in writing the mocks, these steps were implemented by general Python routines, rather than demonstration-specific dictionaries. The local correctness of these steps can be readily evaluated by using the corresponding Python function from the mock as an oracle.

\begin{table*}
\begin{footnotesize}
\begin{center}
\begin{tabular}{llrrr|r}
\hline
 & & \multicolumn{3}{c|}{Single-step prompt accuracy} & \multicolumn{1}{c}{Within trace}\\
 Task & Step & full & $-$ $\mu$-trace & $-$ trace parse & 
 \multicolumn{1}{c}{accuracy} \\
\hline
tracking shuffled objs & simulate\_swap & 96.7 & 73.3 & 100.0 & 100.0 \\
& answer\_question & 100.0 & 66.7 & 100.0 & 100.0 \\
word sorting & kth\_letter & 96.7 & 96.7 & 96.7 & 100.0 \\
 & partition\_words & 100.0 & 100.0 & 96.7 & 100.0 \\
 & sort\_keys & 100.0 & 100.0 & 100.0 & 100.0 \\
& bucket\_sort & 76.7 & 76.7 & 96.7 & 96.7 \\
 & flatten & 100.0 & 100.0 & 100.0 & 100.0 \\
multistep arithmetic  & is\_simple\_expression & 83.3 & 76.7 & 86.7 & 96.7 \\
  & rewrite\_expression & 73.3 & 80.0 & 16.7 & 93.3 \\
  & eval\_simple\_expression & 100.0 & 100.0 & 100.0 & 100.0 \\
  & eval\_expression & 90.0 & 83.3 & 83.3 & 96.7 \\multistep arithmetic  & eval\_variabilized\_expression & 100.0 & 100.0 & 100.0 & 100.0 \\
dyck languages & is\_open\_paren & 100.0 & 100.0 & 100.0 & 100.0 \\
 & update\_stack & 80.0 & 86.7 & 90.0 & 100.0 \\
 & empty\_stack & 86.7 & 90.0 & 80.0 & 100.0 \\
boolean expressions & solve\_negation & 73.3 & 60.0 & 60.0 & 76.7 \\
\hline
Average & & 91.0 & 87.9 & 86.9 & 97.5 \\
\hline
\end{tabular}
\end{center}
\end{footnotesize}
\caption{Testing the accuracy of step-specific prompting, on 16 steps from 5 tasks that can be checked against an oracle.  The "In PTP" column is accuracy of the steps executed in the process of solving task instances in the dev set; Single-step is using the single-step prompting method of Section~\ref{sec:step-by-step}; and the remaining columns are ablations, where micro-traces and trace parsing are not used.}
\label{tab:oracle-details}
\end{table*}

In the BBH suite, there were 16 different steps from five tasks with oracles. We evaluated these oracle-testable steps on the inputs that were actually used in the course of solving the dev set examples (a total of 480 oracle-testable step executions).\footnote{Note that this evaluation can be expensive, since it requires running an LLM many more times than does \ptp{}, as there are an average of 11-12 steps per task.}  Table~\ref{tab:oracle-details} shows the results, as well as macro-averaged overall accuracy. 

The overall performance is more than 90\%, so the steps are  ``meaningful''---i.e., the model ``understands'' then well enough to execute them reliably in a ``step by step'' manner, where each step is performed by an independent LLM call.  Here we use the full method described in Section~\ref{sec:step-by-step}, including the two methods used to address overgeneration, adding micro-traces, and parsing traces to find the first relevant output.  Table~\ref{tab:oracle-details} also shows the result of ablating these, showing that they each contribute to performance.  

The last column of the table also shows the accuracy of the same steps when executed in the context of problem-solving---i.e, the accuracy of the steps when they occur inside a PTP trace.  Note that even the best single-step prompt is significantly worse than performing the corresponding step in the context of problem-solving,\footnote{Overall 10 of the 16 steps have lower average accuracy when executed independently than when executed as part of PTP problem-solving, and the remainder have the same accuracy.  This is significant with $p<0.01$ with a paired $t$-test.} which we discuss further below.

\subsubsection{Why measuring performance of a single step is hard}

We next turn to more general ways of measuring single-step performance.
Unlike the steps in Table~\ref{tab:oracle-details}, in most cases, no oracle for step correctness exists, either because the output contains natural-language text, or because the utility of the step depends on downstream problem-solving in some complex way.  

In general, single-step prompting need not give the same result as executing a step in the context of a larger problem-solving task (as demonstrated by the last column of Table~\ref{tab:oracle-details}). Informally we say that a step is \emph{non-modular} if it behaves differently with a single-step prompt than in the context of a complete trace. The performance of a non-modular step is not easily defined, so we must first address the question of detecting and measuring modularity.  


\begin{table}
\begin{small}
    \begin{tabular}{llrrr}
        Mock Version & Interv. & Acc & Agree & \#Steps\\
        \hline
        baseline             & no & 97.4 & &\\
        $+$ non-mod. step & no & 98.4  & & 4.30 \\
        $+$ non-mod. step & yes & 97.4  & 56.8 & 4.65 \\
        \hline
    \end{tabular}
\end{small}
\caption{Experiments with version of \wctt{sports\_under\-stand\-ing} with a step that has been modified to be non-modular.}
\label{tab:sports-expts}
\end{table}

A simple experiment shows that steps can be highly non-modular. 
We modified the \wctt{consist\-ent\_sports}
step in the \wctt{sports\_under\-stand\-ing} task by removing the first input argument.  Recall that this step compares two sports descriptions, which clearly cannot be done without seeing the first one.  One might expect this prompt to be less accurate, but
evaluated it in the usual way, accuracy is statistically identical to the standard version\footnote{In fact performance is better for the ``broken'' version, which makes 3 errors instead of 5 on the 190 test cases.}, as shown in the first two lines of Table~\ref{tab:sports-expts}. The problem, of course, is that the LLM is not restricted to generate step outputs from the step inputs---in generating step outputs, the LLM can attend to \emph{any} previous text.  

We next explore \emph{forcing} the LLM to consider only the designated inputs to this step, by modifying the trace: we extract the \wctt{consist\-ent\_sports} step, re-execute it with a single-step prompt, and then finally regenerate the remainder of the trace by using the prompting method of Section~\ref{sec:complete-trace}.  Following the terminology of \cite{lanham2023measuring}, we call this process an \emph{intervention}.  Surprisingly, forcing modularity this way still makes no significant different in accuracy! (See the last line of Table~\ref{tab:sports-expts}.)

Closer inspection shows the reason for this. The forced-modular version of the \wctt{consist\-ent\_sports} step does give different outputs---it agrees with the in-context step only 56.8\% of the time. But when \emph{regenerating the remainder} of the trace, the LLM often adds an extra \wctt{consistent\_sports} call, which is then executed in context, and gives a correct answer.  The intervention thus leads to similar accuracy, but results in a slight increase in the  average number of steps.

This suggests detecting non-modularity with statistics in addition to accuracy (e.g., trace length), so as to detect compensations the LLM makes to ``fix'' forced-modular interventions. 

Below we describe this process precisely.


\subsubsection{Defining and measuring modularity} \label{sec:mod-def}

We write the program trace $t^j$ as a series of steps
$$ f_1^j(x_1^j,y_1^j) \ldots f_n^j(x_n^j, y_n^j) a^j
$$
where each $f_i^j$ is a Python function/step name, $x_i^j$ a tuple of inputs, $y_i^j$ an output (also possibly a tuple), and $a^j$ the final answer produced for the initial CoT input $f_1^j,x^j_1$.  For brevity we will drop the superscripts when possible.  

If we model the LLM's generations as a joint distribution over variables $F_1,X_1,Y_1,\ldots,F_n,X_n,Y_n,A$, each $y_i$ depends on the entire preceding sequence, which we write:
\begin{equation} \label{eq:l2r}
    y_i \sim P(Y_i|
 F_{\leq{}i},X_{\leq{}i},Y_{<i})
\end{equation}
where $X_{\leq{}i}$ is shorthand for $X_1,\ldots,X_{i}$, and so on.
Equation~\ref{eq:l2r} holds for any LLM;
a stronger condition is that the $f_i$'s act like actual Python functions, where $y_i$ depends only on $f_i$ and $x_i$:
$$
y_i \sim P(Y_i | F_{i}=f_i,X_{i}=x_i)
$$
In other words, 
each $Y_i$ is conditionally independent of everything else given $X_i$ and $F_i$, i.e.
\begin{equation} \label{eq:indep}
    P(Y_i | F_{i},X_{i}) = P(Y_i | F_{\leq{}i},X_{\leq{}i},Y_{<i})
\end{equation}

We define step $i$ to be \emph{modular} if and only iff Equation~\ref{eq:indep} holds.  
If all steps are modular, then generation is arguably a ``step by step'' process, in the precise sense that the $i$-th output $y_i$ depends only on (1) the choice of what step $f_i$ to use and (2) the selection of the inputs $x_i$ for that step.

\label{sec:mod-measure}

To measure modularity, we 
\emph{force a particular step to be modular}, by replacing its output with the result of a single-step prompt.  After making this change, we then extend the resulting trace prefix using the method of Section~\ref{sec:complete-trace}.  Precisely, if $i$ is the position of the designated step in the trace $t$
$$ t = f_1(x_1,y_1) \ldots f_i(x_i, y_i) \ldots f_n(x_n,y_n) a
$$
then we replace $y_i$ with a new value $\tilde{y}_i$ produced with a single step prompt
$$ \tilde{y_i} \sim P(Y_i | F_i=f_i, X_i=x_i)
$$
and ask the LLM to predict the rest of the trace. This will potentially change every subsequent $x$, $f$, and $y$, yielding the new trace
$$
\tilde{t} = \ldots f_i(x_i, \tilde{y}_i) 
            \tilde{f}_{i+1}(\tilde{x}_{i+1},\tilde{y}_{i+1}) 
\ldots \tilde{f}_n(\tilde{x}_n,\tilde{y}_n) \tilde{a}
$$

Let $P({T}|F_1=f_1,X_1=x_1)$ be the distribution of traces $t$ produced from input $f_1,x_1$, and let $P(\tilde{T}|F_1=f_1,X_1=x_1)$ be the distribution of  traces $\tilde{t}$ after the forced-modularity intervention. Assume that the step of completing a a trace prefix does not change the distribution; 
below we call this assumption the \emph{stable incremental generation assumption.} 
If stable incremental generation holds, and step $i$ is modular, then $P(\tilde{T}|F_1,X_1) = P(T|F_1,X_1)$.  

Similarly, suppose $S$ is any summary statistic derived from a trace, i.e.
$s \sim{} P(S|T)$.
For example, $S$ could be the length of the trace or the correctness of the trace. 
If we assume stable incremental generation, then if step $i$ is modular this also implies 
$$P(\tilde{S}|F_1,X_1) = P(S|F_1,X_1)$$
This means \emph{we can detect failures in modularity if we can detect changes in the distribution of the distributions of the statistics $S$ and $\tilde{S}$}.

While prior work \cite{lanham2023measuring} used accuracy as a summary statistic, we considered additional ones, motivated by Table~\ref{tab:sports-expts}, designed to detect changed behavior in a forced-modular trace.  In addition to $S_\mathrm{corr}$, trace correctness, we consider $S_\mathrm{numSteps}$, the number of steps in the trace; and $S_\mathrm{abTrace}$, the abstract trace of Section~\ref{sec:abstract-trace}.

Specifically, in the experiments below, we first take a sample of $m$ inputs $\mathcal{X}_1=\{x_1^1,\ldots,x_1^m\}$. 
Generally $S$ is just a deterministic function of $T$, so we also write this as $s=g_s(T)$, and compute corresponding samples of the summary statistics 
$\mathcal{S}=\{g_s(t^1, \ldots, g_s(t^m)\}$ and 
$\tilde{\mathcal{S}}=\{g_s(\tilde{t}^1, \ldots, g_s(\tilde{t}^m)\}$.  
We detect changes between $\mathcal{S}$ and $\tilde{\mathcal{S}}$ with a sign test for $S_\mathrm{corr}$,
and a 
paired $t$-test for $S_\mathrm{numSteps}$.  For $S_\mathrm{abTrace}$, we test if the Jensen-Shannon divergence between $S$ and $\tilde{S}$ is higher than expected using a permutation test (see Section~\ref{app:stats}).

\subsubsection{Which tasks are non-modular?} \label{sec:which-nonmod}
\begin{table*}
\begin{footnotesize}
\begin{tabular}{ll|cc|cc|cc|c}
\hline
Task & \multicolumn{1}{l}{Step Name} & \multicolumn{2}{c}{$S_\mathrm{numSteps}$} & \multicolumn{2}{c}{$S_\mathrm{absTrace}$}
 & \multicolumn{2}{c}{$S_\mathrm{corr}$}  & Non \\
 & & $p$ & $\hat{p}$& $p$ & $\hat{p}$& $p$ & $\hat{p}$ & mod?\\
\hline
\multicolumn{2}{l}{\textit{Testing modularity}: force-modularity intervention}  & \\
\hline
boolean expressions & solve\_negation & 0.739 & 0.982 & 0.002 & \textbf{0.014} & 1.000 & 1.000 & \textbf{yes}\\
dyck languages & is\_open\_paren & 0.327 & 0.794 & 0.002 & \textbf{0.014} & 0.500 & 0.992& unclear\\
multistep arithmetic & eval\_simple\_expression & 1.000 & 1.000 & 0.010 & \textbf{0.049} & 1.000 & 1.000 & unclear\\
 & eval\_variabilized\_expression & 1.000 & 1.000 & {0.014} & \textit{0.055} & 1.000 & 1.000 & unclear\\
 & is\_simple\_expression & 0.111 & 0.506 & 0.118 & 0.241 & 1.000 & 1.000 & \\
 & rewrite\_expression & 0.211 & 0.695 & \underline{0.088} & 0.241 & 1.000 & 1.000 & \underline{likely} \\
tracking shuffled objs & simulate\_swap & 0.043 & 0.265 & 0.244 & 0.244 & 1.000 & 1.000 & \\
\hline
\multicolumn{2}{l}{\textit{Testing increm.~gen:} split-and-complete intervention}  &&&&&&& Incr. Gen.?\\
\hline
boolean expressions & solve\_negation & 0.906 && 0.396 && 1.000 & & yes\\
dyck languages & is\_open\_paren &  0.148 && \textbf{0.002} && 0.219 && no\\
multistep arithmetic & eval\_simple\_expression  & 0.169 && \textbf{0.014} && 1.000 && no\\
 & eval\_variabilized\_expression &  \underline{0.083} && 0.242 && 1.000 && \underline{unlikely}\\
 & rewrite\_expression & 1.000 && 0.408 && 1.000 && yes \\
\hline
\end{tabular}
\end{footnotesize}

\caption{Testing non-modularity and incremental generation assumption on oracle-checkable algorithmic tasks.  Unlisted tasks had very few disagreements between the forced-modular step output and the within-PTP step output.  Steps can be assumed to be non-modular if (1) some statistic is significantly different from the no-intervention base, after an intervention in which modularity is forced and (2) there is no evidence that 
generation is not incremental, i.e. no statistic is significantly different after the split-and-complete intervention.
}  \label{tab:oracle-nonmod}
\end{table*}
\begin{table*}
\begin{footnotesize}
\begin{tabular}{ll|cc|cc|cc|c}
\hline
Task & \multicolumn{1}{l}{Step Name} & \multicolumn{2}{c}{$S_\mathrm{numSteps}$} & \multicolumn{2}{c}{$S_\mathrm{absTrace}$}
 & \multicolumn{2}{c}{$S_\mathrm{corr}$}  & Non \\
 & & $p$ & $\hat{p}$& $p$ & $\hat{p}$& $p$ & $\hat{p}$ & mod?\\
\hline
\multicolumn{2}{l}{\textit{Testing modularity}: force-modularity intervention}  & \\
\hline
causal judgement & plausible\_conclusion & 1.000 & 1.000 & 0.020 & 0.345 & 1.000 & 1.000 \\
 & plausible\_inference & 1.000 & 1.000 & 0.024 & 0.369 & 0.500 & 1.000 \\
& relevant\_sentences & 0.489 & 0.999 & 0.651 & 1.000 & 0.500 & 1.000 \\
date understanding & answer\_question & 0.326 & 0.999 & 0.010 & 0.198 & 0.500 & 1.000 \\
 & make\_inference & 0.333 & 0.999 & 0.020 & 0.345 & 1.000 & 1.000 \\
disambiguation qa & find\_possible\_interpretations & 1.000 & 1.000 & 1.000 & 1.000 & 1.000 & 1.000 \\
 & is\_interpretation\_logical & 0.326 & 0.999 & 0.985 & 1.000 & 0.453 & 1.000 \\
formal fallacies & prove & 0.595 & 1.000 & 0.002 & \underline{0.051} & 1.000 & 1.000 & unclear\\
 & to\_logical\_form & 0.063 & 0.763 & 0.192 & 0.967 & 0.375 & 1.000 \\
hyperbaton & classify\_adjective & 0.043 & 0.655 & 0.008 & 0.169 & 1.000 & 1.000 \\
movie recommendation & explain\_best\_choice & 1.000 & 1.000 & 0.464 & 1.000 & 0.500 & 1.000 \\
 & movie\_properties & 0.043 & 0.655 & 0.244 & 0.982 & 1.000 & 1.000 \\
 & summarize\_movies & 1.000 & 1.000 & 1.000 & 1.000 & 1.000 & 1.000 \\
penguins in a table & answer\_question & 0.161 & 0.975 & 0.234 & 0.982 & 0.250 & 0.999 \\
 & table\_operation & 0.165 & 0.975 & 1.000 & 1.000 & 1.000 & 1.000 \\
reasoning @ colored objs & query\_colored\_objects & 1.000 & 1.000 & 0.813 & 1.000 & 1.000 & 1.000 \\
ruin names & humorous\_edit & 0.161 & 0.975 & 0.176 & 0.963 & 1.000 & 1.000 \\
 & meaningful\_edit & 0.184 & 0.975 & 0.154 & 0.951 & 0.500 & 1.000 \\
salient translation & choose\_answer & 0.326 & 0.999 & 1.000 & 1.000 & 1.000 & 1.000 \\
 & choose\_error\_type & {0.000} & \textbf{0.003} & {0.002} & \underline{0.051} & 1.000 & 1.000 & \textbf{yes}\\
 &find\_translation\_error & {0.000} & \textbf{0.000} & 0.002 & \underline{0.051} & 1.000 & 1.000 & \textbf{yes}\\
snarks & is\_sarcastic & 1.000 & 1.000 & 1.000 & 1.000 & 1.000 & 1.000 \\
 & judge\_statement & 1.000 & 1.000 & 1.000 & 1.000 & 0.625 & 1.000 \\
 & summarize\_statement & 1.000 & 1.000 & 1.000 & 1.000 & 0.031 & 0.562 \\
sports understanding & consistent\_sports & 0.326 & 0.999 & 0.805 & 1.000 & 1.000 & 1.000 \\
 & sport\_for & 0.326 & 0.999 & 0.695 & 1.000 & 1.000 & 1.000 \\

\hline
\multicolumn{2}{l}{\textit{Testing increm.~gen:} split-and-complete intervention}    &&&&&&& Incr. Gen.? \\
\hline
formal fallacies & prove & \underline{0.082} && 0.412 && 0.625 && perhaps\\
salient translation&  choose\_error\_type  & 0.326 && 1.000 && 1.000 && yes \\
salient translation& find\_translation\_error & 1.000 && 1.000 && 1.000 && yes
\\
\hline
\end{tabular}
\end{footnotesize}

\caption{Testing non-modularity and incremental generation assumption on selected NLP tasks.  Unlisted tasks had very few disagreements between the forced-modular step output and the within-PTP step output.  Steps can be assumed to be non-modular if (1) some statistic is significantly different from the no-intervention base, after an intervention in which modularity is forced and (2) there is no evidence that 
generation is not incremental, i.e. no statistic is significantly different after the split-and-complete intervention.
}  \label{tab:nlp-nonmod}
\end{table*}

Let's briefly review the preceding subsections.

In Section~\ref{sec:oracle} we showed that we can prompt the model to accurately execute individual steps in isolation, using the method of Section~\ref{sec:step-by-step}.
    
In Section~\ref{sec:mod-def}, we defined and motivated modularity: steps are \emph{modular} if they depend only on their defined inputs, and not other aspects of the trace in which they are embedded.  We then proposed a way to measure modularity of a step by using forced-modularity interventions.  
A subtlety is that this intervention is based on two changes: (a) replacing the step output with a forced-modular version, and then (b) re-generating the rest of the trace.  Hence we can only measure the impact of distributional changes due to (a) if we are confident that (b) does not also change the distribution, as assumption we call \emph{stable incremental generation.}  

In the top of Table~\ref{tab:oracle-nonmod}, we evaluate the result of the forced-modularity intervention on some of the oracle-checkable steps\footnote{We do not use steps that call other steps in this experiment.}  from Table~\ref{tab:oracle-details} on the dev data.    The columns labeled $p$ are the $p$-values from our permutation test, so a small value indicates that the step changed the distribution of the statistic.  The column $\hat{p}$ is the $p$-value after a multiple-test correction.\footnote{If you perform many tests without this correction, you can expect some non-significant differences to have low $p$-values by chance.  The multiple-test correction adjusts for this.}  Corrected $p$-values below 0.1 are italicised and corrected $p$-values values below 0.05 are boldfaced: these correspond to steps where some statistic is changed by the forced-modularity intervention.

For steps where corrected $p$-values are low, we also tested the assumption of stable incremental generation with another intervention, where only (b) occurs.  We call this intervention split-and-generate: we  truncate the trace at step $i$ and then re-generate it.  We see that the assumption fails for three of the steps, so we cannot conclude that these steps are non-modular; hence the final column is labeled ``unclear'' for them.  There is no evidence against the stable incremental generation assumption for the ``solve negations'' step, but weak evidence ($p=0.083$) against it for ``rewite expressions'' step, so we mark the boolean expressions step (with $\hat{p}=0.014$ and stable incremental generation) as non-modular and the other (with $\hat{p}=0.083$ and stable incremental generation) as ``likely'' to be non-modular.

We note that the nonmodular steps are all cases where the single-step prompts underperformed in Table~\ref{tab:oracle-details}.  We also note that correctness is the \emph{least} useful summary statistic on this data.

The situation is quite different for the algorithmically simpler NLP tasks.  In Table~\ref{tab:nlp-nonmod} we summarize the results of repeating this experiment on 25 steps from 12 NLP tasks. (We choose steps which were prone to local errors as being relatively more interesting.)  Only 3 of these steps show a significant distributional shift with the forced-modularity intervention, and one of these may be explained by a failure of the stable incremental generation assumption.

We conclude that non-modularity is rare in the NLP tasks.  We note again that accuracy, the traditional statistic for interventions, is not sensitive enough to detect any changes here.

\subsubsection{Discussion}

These results show that \emph{on non-algorithmic tasks, non-modularity is rare}.  In fact, of the 12 NLP tasks we consider, only one contained any steps that appear to be non-modular.    

This observation, of course, depends in part on the specific mocks that we used in these experiments.  Another disadvantage of this test is that, while one can identify non-modularity (when a null hypothesis is refuted), you cannot ensure modularity---the null might still be refuted if you examine more data.  However, it does seem clear that at least most step executions of a step are modular.


\section{Related Work}

\subsection{Understanding CoT and Measuring Faithfulness}

There has been a great deal of work on CoT prompting and reasoning in LLMs, as surveyed here \cite{huang2022towards}. 
The prior work that is most relevant to this paper is analytic work on understanding why CoT works, and why and when it fails.  A number of hypotheses have been proposed for CoT's success: for instance, 
it has been argued that the longer outputs of CoT provide additional computational power, which is needed to solve computationally complex tasks \cite{feng2024towards}. More notably, it has been argued that CoT ``unlocks'' reasoning abilities that emerge in sufficiently large language models \cite{wei2022chain,prystawski2024think}. This view is supported by the observation that phrases like ``let's think step by step'' often lead to improved LLM performance on many tasks \cite{kojima2022large}.  

The metaphor of ``unlocking'' reasoning is very compelling, and may be why there has been little prior research on the degree to which CoT prompts \emph{teach} LLMs task-specific reasoning methods, and the degree to which LLM behavior can be controlled by presenting more explicit method traces (as we showed is possible using \ptp).
The results of Sections~\ref{sec:error-types} and \ref{sec:why-nonlocal} illustrate that CoT prompts also provide function by providing new information to the LLM, as its steps describe a particular \emph{strategy} for  a task.

The notions of modularity and locality explored in this paper are closely related to the notion of CoT explanation \emph{faithfulness} \cite{jacovi-goldberg-2020-towards, turpin2024language, lanham2023measuring}.  Turpin \textit{et al} noted that when LLMs are biased (e.g., illustrating a gender bias in predictions of occupation), CoT prompting sometimes leads to predictions that preserve the underlying bias, but still produces plausible-looking CoT explanations---explanations that do \emph{not} reflect the underlying bias.  This sort of ``unfaithful'' explanation is disturbing, perhaps because it seems like deception; practically it is also inconvenient if explanations are used to justify LLM predictions to a user, or even if explanations are used to help optimize LLM behavior.  Later work (e.g., \cite{bao2024llms}) has reinforced these observations, and various proposals have been made to measure faithfulness; some even adopt very general approaches that numerically measure influence in neural networks \cite{parcalabescu2024measuringfaithfulnessselfconsistencynatural} to address perceived limitations in measuring faithfulness \cite{bentham2024chainofthoughtunfaithfulnessdisguisedaccuracy}.

Here we build on the work of \cite{lanham2023measuring}, where faithfulness is measured by changing a generated explanation (e.g., introducing errors, paraphrasing, etc) and then measuring the impact on accuracy.  We extend this work by considering a particular change (forced modularity of a step); using new statistics to measure the impact of a change; and studying the more precise notions of modularity of steps and error locality. Our work is also influenced by \cite{yee2024faithful}, who explore error recovery in CoT models and its effect on measuring faithfulness.

\subsection{Calling Python from LLMs}

One method that has been proposed to improve faithfulness, and more generally to enhance reasoning abilities, is to combine LLMs with symbolic methods--often methods based on Python.  For example, in Program of Thoughts \cite{chen2022program} and Program Aided LMs \cite{gao2023pal} and elsewhere \cite{lyu2023faithfulchainofthoughtreasoning}, an LLM generates a Python program for each task instance, which is then executed to produce an answer.  An alternative to this architecture is tool use \cite{paranjape2023art}, where an LLM is augmented interleave text generation and execution of symbolic tools.
A third architectural alternative is to generate Python-like ``pseudocode'' \cite{li2023chain, weir2024learning,chae2024language}---code which has a symbolic, Python-like syntax, but includes routines that cannot be executed directly, but must be emulated by an LLM,  or Python code which uses LLMs as a tool \cite{zhou2024conceptualunbiasedreasoninglanguage}.  

Of the approaches above, the ``pseudocode'' architecture is closest to \ptp{}.  To our knowledge, however, prior research in this direction have sought to produce a single program that works for all instances of a task.  In contrast, our approach does not produce a program, or even require that such a program exists, since a trace is generated for each task instance.

A second difference is that unlike the prior work discussed above, \ptp{} does not use a hybrid LLM/symbolic architecture, but is a pure prompting method, thus providing an improvement in simplicity. We also quantify and measure faithfulness in a novel way.

\subsection{Calling LLMs from Python}

Another widely-used methodology for developing applications with LLMs is to manually decompose a task into substeps that can be performed by making a series of calls to LLMs.  Examples of software systems that support this methodology include DsPy \cite{khattab2023dspy} and LangChain \cite{topsakal2023creating}.  Like \ptp{}, this approach allows one to cleanly decompose a task, and measure and optimize performance on the subtasks.  \ptp{} performs this sort of decomposition within a single prompt, which will generally be more efficient.

\subsection{CoT Prompt generation}

CoT prompting is powerful, but it is expensive to collect examples, and difficult to optimize CoT prompts,
although these problems can been ameliorated by synthetic generation of CoT prompts \cite{pmlr-v202-shao23a} and prompt optimization \cite{wang2023promptagentstrategicplanninglanguage}.  Prior work has also explored semi-automatic CoT prompt generation by making use of post-hoc extractive rationales of classification decisions \cite{krishna2024post}.

\ptp{} proposes new ways to semi-automatically produce CoT prompts; however unlike prior work, these prompt-generation approaches are fairly specific to the generation of PT prompts. 

\subsection{Learning programs from traces}

Learning from program traces has been studied for many decades in machine learning (e.g., \cite{smith1984synthesis}), and has gone by various names including programming by example and learning by demonstration \cite{cypher1993watch,lieberman2001your,lau2003learning}.  To our knowledge, there has been little work on this topic in the LLM era, although  algorithmically produced programs traces have been used to analyze the noise-sensitivity of CoT reasoning
\cite{havrilla2024understandingeffectnoisellm}.  Our work differs from \cite{havrilla2024understandingeffectnoisellm} in that we propose program traces as a way of constructing controllable chain-of-thought prompts for end tasks, rather than as a tool for generation of analytic data.

\section{Conclusions}

We propose \ptp{}, a variant of CoT prompting in which few-shot
CoT demonstrations are wrapped in a semi-formal
syntax which (1) identifies and names steps; (2) defines the input/output behavior of steps; and (3) replaces CoT explanations of in-context examples
with chains of these formalized steps on the same
example.  We show that this approach is broadly applicable, in that comparable accuracies can be obtained by many models on many tasks.  It also offers some advantages over prior prompting schemes.

First, \ptp{} traces are \emph{easier to analyze automatically than arbitrary CoT explanations}, since they follow the syntax used for steps---even maintaining the type-correctness of step inputs and outputs, in most cases.  This means we can extract inputs and outputs of individual steps, count the number of inference steps, and perform other types of analysis easily.  For example, we can more easily localize errors in an incorrect explanation.  We can also identify errors which cannot be localized to steps (e.g., copying errors where the input to one step should be output of another, but is not).  Above we showed that \emph{trace entropy} correlates with the presence of non-local errors: simply put, more complicated reasoning strategies are harder to learn from CoT prompts than simpler ones.

Second, \ptp{} allows prompts to \emph{elicit results of executing a single step of a CoT process in isolation}, outside of a larger reasoning task.  This enables a new class of interventions, where we force a step inside a trace to be \emph{modular} by executing it in isolation, as if it were performed in by an agentic LLM program.  By doing this and measuring statistical perturbations in the result, we can detect if the step actually depends only on its declared inputs, or if it depends on other previous steps in unexpected ways.  We show that this sort of non-modular behavior can occur, but it is rare in our collection of tasks.

\section*{Acknowledgments}

Thanks to Chenyan Xiong, Kelvin Gu, Fernando Pereira, and many other colleagues for comments on early versions of this work.

\bibliography{main}
\bibliographystyle{acl_natbib}



\appendix

\section{Details on Methods} 

\subsection{CoT Prompts}\label{app:cot}

The original CoT prompts from \cite{suzgun2022challenging} are designed to give an explanation, followed by a short answer which can be checked against a target.  Modern RLHF- and instruction- trained models do not always follow the syntax of CoT examples, so we modified the prompts as suggested by the example below, where the underlined text was added:

\begin{small}
\begin{quotation}
Determine whether an artificially constructed sentence relating to sports is plausible or not.

\smallskip

\textit{When you give your answer, follow the format of the examples below
carefully.  In particular, you MUST end your answer with either 'Final
answer: yes' or 'Final answer: no'.}

\smallskip

Q: Is the following sentence plausible? "Bam Adebayo scored a reverse layup in the Western Conference Finals."

\smallskip

A: Let's think step by step. Bam Adebayo is an American basketball player. Scoring a reverse layup in the Western Conference Finals is part of the NBA Finals. So the answer is yes.\\
\textit{Final answer: yes}
\end{quotation}
\end{small}

For multiple-choice questions, the new instruction says ``\textit{\ldots you MUST end your answer with the line 'Final answer: (X)' for some letter X.}''

In addition to making it easier to recognize the answer, the instruction to \textit{follow the format of the examples carefully} also encourages CoT reasoning and makes the CoT baseline more comparable to the PTP method.

Because robustness to noisy examples is not the focus of this paper, we also corrected two errors in the CoT examples: a set of incorrect coordinates in \texttt{navigate}, where positive and negative examples were swapped, and one movie name in \texttt{movie recommendation} which was broken across two options (``They Shoot Horses Don't They'' was split into two movie choices, ``(A) They Shoot Horses'' and ``(B) Don't They''.)
The prompts used are available in \ourgh{} under \texttt{modified-cot-prompts}.

\subsection{Generation of traces} \label{app:tracegen}

\begin{figure*}[t]
\begin{minipage}[c]{0.45\textwidth}
    \includegraphics[width=1.0\textwidth]{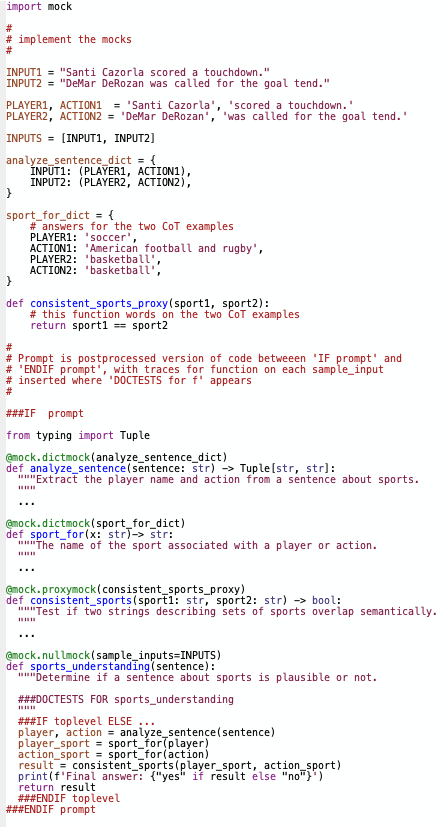}
\end{minipage}~\begin{minipage}[c]{0.45\textwidth}
    \includegraphics[width=1.0\textwidth]{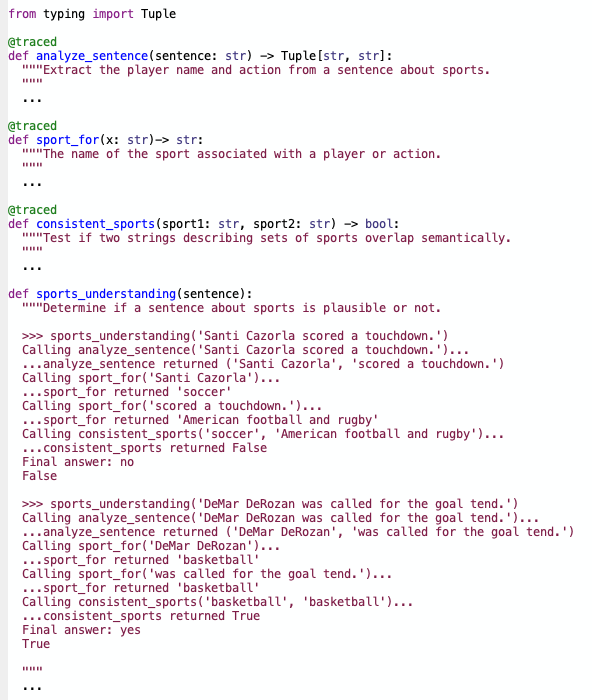}
\end{minipage}
\caption{Left, code for a mock for a simplified HB task. Right, a Program Trace prompt derived from the mock.}  \label{fig:mock}
\end{figure*}

In our implementation, mocks are special Python programs, and invoking a mock program from the command-line with appropriate arguments will produce the ``program'' shown in Step (1) of Figure~\ref{fig:fig1}.  Mock programs can include preprocessing directives---e.g., code that is not surrounded by the lines \verb|###|\wctt{IF prompt} and \verb|###|\wctt{ENDIF prompt [ELSE ...]} will not be included in the Step (1) program.  Other preprocessing directors specify where to insert generated traces (e.g., \verb|###|\wctt{DOCTESTS FOR foo}).  The ``stub'' for a step can be linked to a mock implementation by decorators like \wctt{@dictmock} for a dictionary-based implementation and \wctt{@proxymock} for a Python function implementation. A similar decorator associates a top-level function with the inputs used to automatically generate the traces/demonstrations from the mock.  Figure~\ref{fig:mock} shows an example of a simple mock and the corresponding generation.

\subsection{Prompts for generated traces and continuing partial traces} \label{app:continue-trace}

The prompts to generate a trace, and to continue a partial trace, are shown in Figure~\ref{fig:gentrace}.

\begin{figure*}[t]
\begin{minipage}[c]{0.45\textwidth}
    \includegraphics[width=1.0\textwidth]{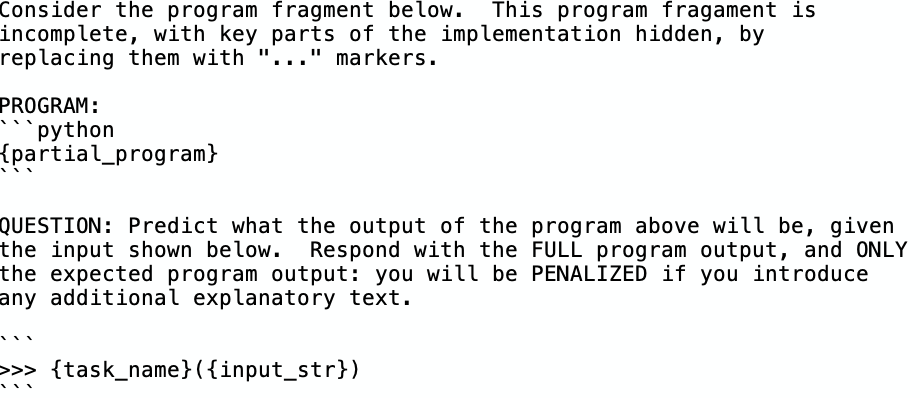}
\end{minipage}~\begin{minipage}[c]{0.45\textwidth}
    \includegraphics[width=1.0\textwidth]{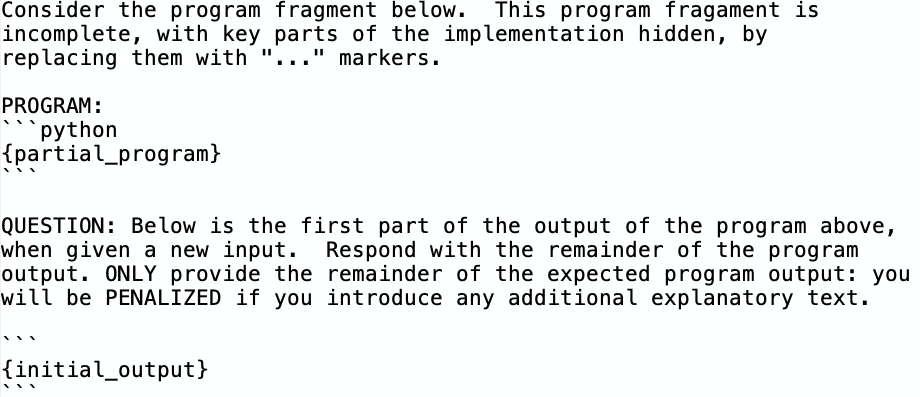}
\end{minipage}
\caption{Prompts to generate a trace and continue a partial trace.}  \label{fig:gentrace}
\end{figure*}

\subsection{Single-step prompts} \label{app:single-step}

The standard prompt to generate a trace was used, but we address over-generation in two ways, as discussed.
\begin{itemize}
\item Trace parsing: To extract the output of the step $f_i$, we analyze the trace to find all pairs of matching lines of the form \wctt{Calling $f_i$($x_i$)...} and \wctt{... $f_i$ returned $y_i$}.  We then use the string $y_i$ for the first of these pairs as the output.
\item Micro-traces: For the steps $f_i$ for which we plan to perform single-step prompting, we insert into the stub for $f_i$ the preprocessing directive \verb|###|\wctt{DOCTESTS FOR $f_i$ IMPLIED BY $h$}, where $h$ is the name of the top-level mock function (and also the name of the task.)  At program generation time, this directive trace-parses the generated traces for $h$, the top-level function, and detects all of the \wctt{Calling}/\wctt{returned} line pairs for $f_i$.  The first $K=3$ of these pairs (together with all lines between a pair) are inserted in place of the preprocessing directive.
\end{itemize} 

\section{Experimental details} \label{app:expts}

\subsection{Avoiding overfitting to the test data}

To avoid overfitting, only one PT prompt was evaluated on the test set in Table~\ref{tab:ptp-vs-cot}, except for three cases in which problems discovered in the PTP prompt late in experimental procedure. For these the PT prompt result of Table~\ref{tab:ptp-vs-cot} is the second prompt evaluated. 

One case (Navigation) had a bug in which intermediate results were not printed in some cases.  In two cases (Date Understanding and Temporal Sequences) review of the mock algorithm determined it was quite different from a CoT-suggested method, and the PT prompt was replaced with one following a more similar strategy. Recall the goal of this paper is exploring new ways to observe and analyze CoT explanations, not to explore performance differences on these specific tasks; hence we want to align the algorithms implied by the PT prompts with those implied by the CoT prompts as much as possible.  

\subsection{Comparing CoT and PTP}

In the experiment of Table~\ref{tab:ptp-vs-cot}, it is possible for models to consistently fail on an example.  The most common reasons for this are that the LLM's ``guard rails'' are triggered and it refuses to produce output (e.g., some of the Causal Reasoning examples involve questions about bombs and shootings); that that model's RLHF instruction-training makes it reluctant to follow the prompt (the Salient Translation Errors suffered from this); or that the model's output was too long\footnote{Model output limits are trivial to get around by generating output in stages, but we leave implementation of this for later work.} (which is more frequently a problem for PTP, which has bulkier prompts and outputs.) The statistics above are computed for the subset of examples for which neither model was blocked.  For Salient Translation, we added one additional sentence\footnote{``Do NOT simply answer the question with a multiple-choice answer, always generate the program trace first.''} to the prompt template used in Stage 3 to improve coverage.  The number of examples blocked was moderate for every task, never more than 15\%.

\begin{table*}[tbh]
\begin{small}\begin{center}
\begin{tabular}{lrrr|c}
\hline
 & & CoT  &  & Chain-of-Code\\
  & ours & da-vinci$^1$ & codex$^2$ & da-vinci $^1$\\
\hline
Average NLP     & {84.0} & 69.7 & 73.5 & 72.6 \\
Average Alg     & 88.9   & 74.1 & 74.4 & 94.7 \\
Average Overall & {86.4} & 72.0 & 73.9 & 83.2 \\
\hline
\end{tabular}
\end{center}\end{small}
\caption{Overall results of Table~\ref{tab:ptp-vs-cot} compared with prior work: results (1) are from \cite{li2023chain}, (2) are from \cite{suzgun2022challenging}.  Prior results are over the full BBH test sets, whereas our results are on a sample of approximately 75\%, as described in the text.
The chain-of-code method is not a single prompt, and executes some code with Python, so is not directly comparable.}
\label{tab:prior}
\end{table*}

\subsection{Assessing type correctness}

We used the \wctt{typeguard} package to verify type correctness.

To test syntactic correctness, we could usually simply call Python's \wctt{eval} function of LLM-generated text.  However, in one mock, we used a library class (\wctt{intEnum}), which has a printed representation that cannot be simply read and \wctt{eval}ed.  To verify correctness for this we wrote a custom parser for \wctt{intEnum} objects that de-serializes the printed representation (which looks something like \verb|<AdjectiveCategory.COLOR: 5>|).

\subsection{Steps used for testing modularity} \label{app:oracle-expts}

The tasks used for testing modularity are listed below.

NLP tasks:
\begin{itemize}
\item causal judgement:  relevant sentences, plausible inference, plausible conclusion
\item date understanding:  make inference, answer question
\item disambiguation qa:  find possible interpretations, is interpretation logical
\item formal fallacies:  to logical form, prove
\item hyperbaton:  classify adjective
\item movie recommendation:  movie properties, summarize movies, explain best choice
\item penguins in a table:  table operation, answer question
\item reasoning about colored objects:  query colored objects
\item ruin names:  meaningful edit, humorous edit, first edit is more humorous proxy
\item salient translation error detection:  find translation error, choose error type, choose answer
\item snarks:  summarize statement, judge statement, is sarcastic
\item sports understanding:  sport for, consistent sports
\end{itemize}

Algorithmic tasks (all of these are also oracle tasks):
\begin{itemize}
\item boolean expressions:  solve negation
\item dyck languages:  is open paren
\item multistep arithmetic two:  is simple expression, eval simple expression, eval variabilized expression, rewrite expression
\item tracking shuffled objects three objects:  simulate swap, answer question
\item word sorting:  flatten, kth letter, sort keys
\end{itemize}

\subsection{Statistical tests} \label{app:stats}

The sign test was implemented by testing a binomial estimating 
$P(\tilde{C}=1 | \tilde{C} \not= C)$,
where $C$ is binary correctness in the original trace and
$\tilde(C)$ is correctness in the forced-modular trace.
Non-modularity is detected if this estimate is different from $0.5$ with confidence at least 95\%, using \wctt{scipy.stats.binomtest} (which is an exact test).

The permutation tests uses the default settings for 
\wctt{scipy.stats.permutation test}, except for only 1000 samples (and a two-sided test).
Jensen-Shannon divergence (which is the average KL-divergence of each sample distribution to the mean of the two distributions) was computed using \wctt{scipy.spatial.distance.jensenshannon}.

Corrections for multiple tests were done with the Bonferroni correction as implemented by \wctt{statsmodels.stats.multitest.multipletests}.

\end{document}